\documentclass[sigconf]{acmart}

\AtBeginDocument{%
  }

\copyrightyear{2024}
\acmYear{2024}
\setcopyright{rightsretained}
\acmConference[RecSys '24]{18th ACM Conference on Recommender Systems}{October 14--18, 2024}{Bari, Italy}
\acmBooktitle{18th ACM Conference on Recommender Systems (RecSys '24), October 14--18, 2024, Bari, Italy}\acmDOI{10.1145/3640457.3688054}
\acmISBN{979-8-4007-0505-2/24/10}

\usepackage{bbm}

\begin{document}

\title{Privacy Preserving Conversion Modeling in Data Clean Room}



\author{Kungang Li}
\affiliation{%
  \institution{Pinterest Inc.}
  \country{USA}
}
\email{kungangli@pinterest.com}

\author{Xiangyi Chen}
\affiliation{%
  \institution{Pinterest Inc.}
  \country{USA}
  }
\email{xiangyichen@pinterest.com}

\author{Ling Leng}
\affiliation{%
  \institution{Pinterest Inc.}
  \country{USA}
}
\email{lleng@pinterest.com}

\author{Jiajing Xu}
\affiliation{%
 \institution{Pinterest Inc.}
 \country{USA}
 }
\email{jiajing@pinterest.com}

\author{Jiankai Sun}
\affiliation{%
  \institution{ Pinterest Inc.}
  \country{USA}
  }
\email{jiankaisun@pinterest.com}

\author{Behnam Rezaei}
\authornote{Currently at Roblox. Work was done while the author was employed at Pinterest Inc..}
\affiliation{%
  \institution{Pinterest}
  \country{USA}
  }
\email{brezaei@pinterest.com}



\renewcommand{\shortauthors}{Kungang Li et al.}

\begin{abstract}
In the realm of online advertising, accurately predicting the conversion rate (CVR) is crucial for enhancing advertising efficiency and user satisfaction. This paper addresses the challenge of CVR prediction while adhering to user privacy preferences and advertiser requirements. Traditional methods face obstacles such as the reluctance of advertisers to share sensitive conversion data and the limitations of model training in secure environments like data clean rooms. We propose a novel model training framework that enables collaborative model training without sharing sample-level gradients with the advertising platform. Our approach introduces several innovative components: (1) utilizing batch-level aggregated gradients instead of sample-level gradients to minimize privacy risks; (2) applying adapter-based parameter-efficient fine-tuning and gradient compression to reduce communication costs; and (3) employing de-biasing techniques to train the model under label differential privacy, thereby maintaining accuracy despite privacy-enhanced label perturbations. Our experimental results, conducted on industrial datasets, demonstrate that our method achieves competitive ROC-AUC performance while significantly decreasing communication overhead and complying with both advertisers' privacy requirements and user privacy choices. This framework establishes a new standard for privacy-preserving, high-performance CVR prediction in the digital advertising landscape.

\end{abstract}






\maketitle

\section{Introduction}
In online advertising systems, conversion serves as a tangible measure of user satisfaction and represents a definitive indication of success in an e-commerce environment \cite{ma2018entire, chapelle2014modeling, pan2019predicting, lee2012estimating}. It is often the primary objective for advertisers in numerous marketing contexts. Estimating the conversion rate (CVR) is essential in performance-based digital advertising \cite{guo2022calibrated, haramaty2023extended, xu2022ukd}.

However, some advertisers hesitate to share conversion data with the advertising platforms in order to protect their user conversion data privacy and maintain competitive advantage. Clean room is an option they are interested in sharing the conversion data in a more privacy centric way. A data clean room is a secure environment where organizations can collaborate and share data without exposing sensitive information to unauthorized parties \cite{herbrich2022data, johnson2022privacy}. They are viewed as a trusted third party for both parties, therefore, matching conversion data and user data from both parties in the clean room is considered as an viable option for conversion modeling. But nowadays, clean rooms may have limited computing capability, limited training visibility and high pricing structure thus it is not feasible to train complicated models in the clean room.

To overcome the limitations, we borrow the idea of split learning, which aims at collaboratively training a model using the private input and label data held by two separate parties \cite{liu2024vertical, tran2022privacy, li2021label}. We have identified two primary reasons why the standard split learning method is not suitable for our scenario. Firstly, there are privacy concerns associated with standard split learning. This arises from the fact that the label information can potentially be inferred by adversaries through shared sample-level gradients \cite{li2021label, fu2022label, yang2022dp}, which are exchanged between the clean room and the advertising platform. These are significant concerns for advertisers, thereby making the adoption of standard split learning infeasible for us. Secondly, the inference process in the advertising platform necessitates the complete model, including the layers within the clean room. This requirement diverges from the standard split learning approach, where only a portion of the model is used for inference.

In this work, we proposed an approach to train large conversion models with a new framework. The approach employs parameter efficient fine-tuning techniques, differential privacy, and communication-efficient training algorithms to simultaneously comply with privacy restrictions for conversion data and enable efficient coordination between clean room and advertising platforms to train CVR models.

\section{Methodology}
\label{sec:methods}

\subsection{Model training framework}

\begin{figure}[t]
\includegraphics[width=1.0\linewidth]{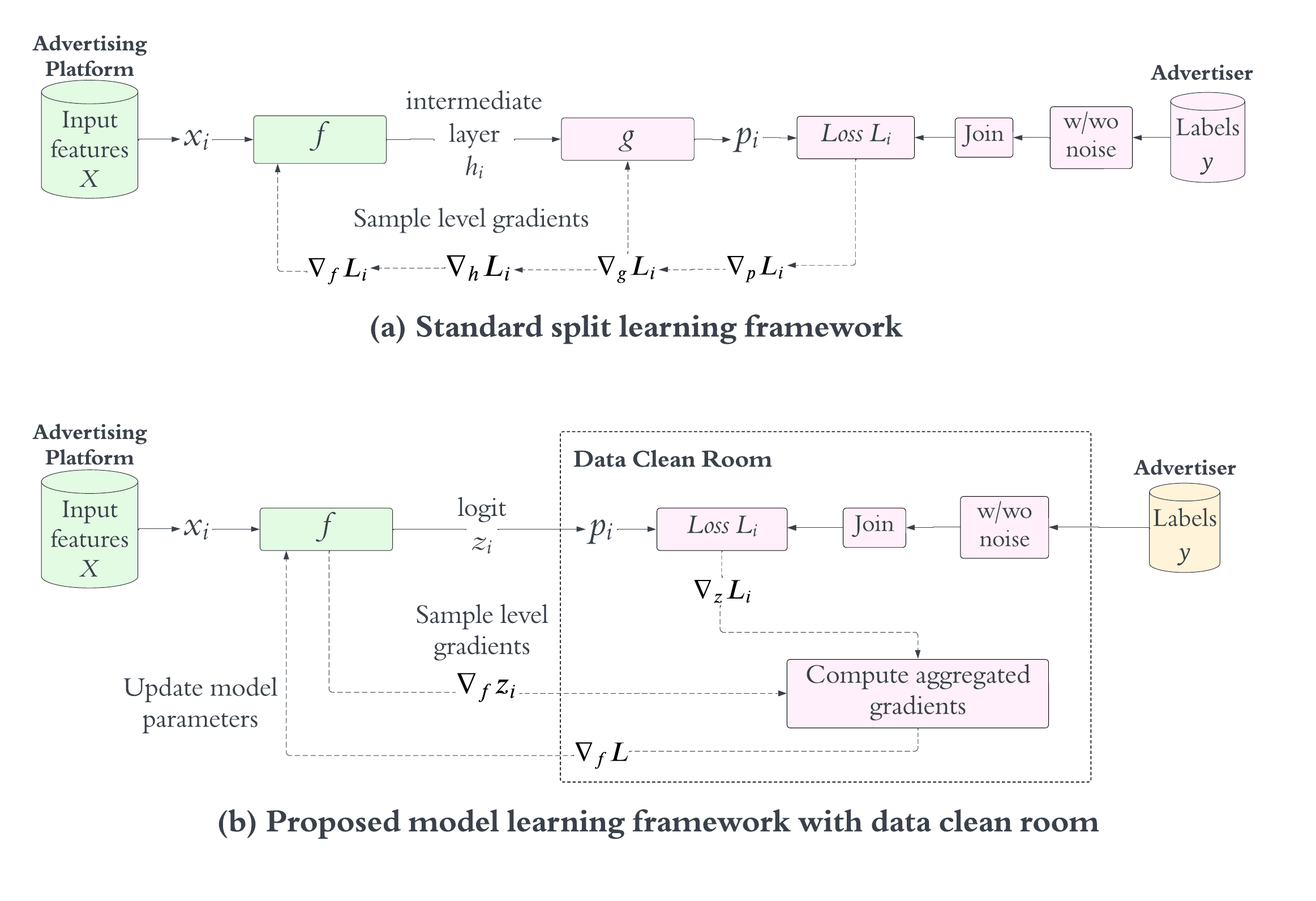}
\caption{Comparison of (a) standard split learning framework and (b) the proposed framework to train conversion models in the data clean room.}
\label{fig:proposed_framework}
\Description[Proposed framework to train conversion models in the data clean room with aggregated gradients]{The feature party conducts most computation of the forward pass and send the computed logits to the clean room, which are joined with labels to compute the training loss. To compute the gradients with respect to the model parameters without disclosing sample level gradients from the clean room to the advertising platform, the aggregated gradient is computed in the clean room, and are returned to the advertising platform to update model parameters.}
\end{figure}

Figure \ref{fig:proposed_framework} compares the standard split learning and our proposed approach with the context that the advertising platform is the party with raw input features (the feature party) and the advertiser is the party with ground truth labels (the label party). In our proposed framework, the feature party conducts most computation of the forward pass and send the computed logits to the clean room, which are joined with labels to compute the training loss. To compute the gradients with respect to the model parameters without disclosing sample level gradients from the clean room to the advertising platform, the aggregated gradient is computed in the clean room. The feature party computes and sends the partial derivatives of the logits with respect to the model parameters $\frac{\partial z_i}{\partial f}$, which are joined with the loss with respect to logit partial derivatives $\frac{\partial L_i}{\partial z_i}$ to compute $\nabla_f L$ \footnote{We do not need to conduct other complicated computation within the clean room.}. In our case, $\nabla_f L$ is given by $\sum_{i=1}^b (p_i - y_i) \frac{\partial z_i}{\partial f}$, where binary cross-entropy loss with sum reduction is applied, $p_i$ is the prediction score for being positive for sample $i$, $y_i$ is the ground truth label, and $b$ denotes the batch size.  The aggregated gradients are returned to the advertising platform to update model parameters. Because we use very large batch size (tens of thousands) during training, the label information from the aggregated gradients shall not be easily attacked by the gradient matching technique such as DLG \cite{zhu2019deep}. In order to strengthen privacy protection, we integrate differential privacy, specifically when the number of trainable parameters $\|f\|$ surpasses the batch size $b$. This strategy effectively defends against potential adversaries who attempt to solve the aforementioned equation, where there are $b$ unknown labels, by extracting label information from a set of equations formed by the number of parameters $\|f\|$. 

\subsection{Adapter based efficient training}

\begin{figure}[H]
\centering
\includegraphics[width=0.75\linewidth]{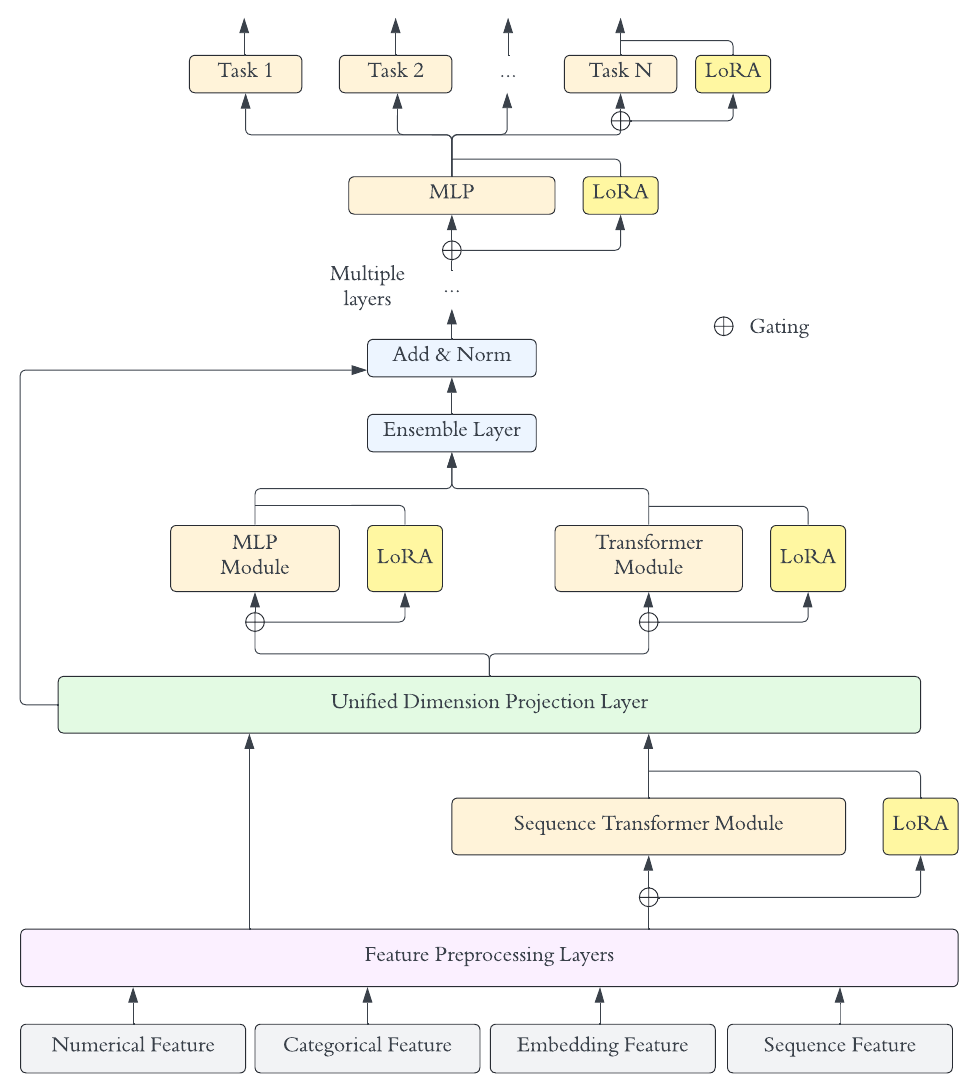}
\caption{Adapter based CVR model.}
\label{fig:proposed_fine_tuning_framework}
\Description[CVR model with LoRA adapters]{Adding gated LoRA adapter layers to the CVR model: the selected layers include the QKV layer in multi-head attention in the sequence transformers, the linear layers in the feature interaction modules, and the linear layers in the multi-task towers.}
\end{figure}

In modern ads delivery systems, the CVR models used in late stage ads ranking are usually very large, with model sizes from a few \textit{GBs} to \textit{TBs} \cite{zhang2022dhen, zhang2024wukong, zhai2024actions, xia2023transact}. 
This creates a significant challenge due to the high communication costs associated with transmitting sample-level partial derivatives from the advertising platform to the clean room. In our use case, it would take about $2.5$ days to complete training on $1$ day's data for some big advertisers with $10$ \textit{Gbps} network bandwidth, which makes it impractical to land into production. To reduce the communication cost, we propose two approaches: (1) Reduce the trainable model parameters by applying pretrain and parameter efficient fine tuning approach; (2) Compress the gradients. The per-sample partial derivatives are compressed before sending to clean room. 

Inspired by adapter approach in large language model (LLM), such as Low-Rank Adaptation (LoRA) \cite{hu2021lora}, we proposed a gated adapter approach for this conversion modeling use case. We apply plug-in adapter module with some trainable parameters to selected layers of the CVR model during fine tuning.

We add gated LoRA adapter layers to the core modules of the CVR model, include the QKV layer in multi-head attention in the sequence transformers, the linear layers in the feature interaction modules, and the linear layers in the multi-task towers (Figure \ref{fig:proposed_fine_tuning_framework} ).

Depending on the model size and network bandwidth between the advertising platform and the clean room, communication cost may need to be further reduced. In this case, the advertising platform can employ gradient compression techniques \cite{xu2021grace} like QSGD \cite{alistarh2017qsgd}, top-k \cite{aji2017sparse}, and powerSGD \cite{vogels2019powersgd} to compress the per-sample partial derivatives sent to the clean room. 

\subsection{Training with label DP}

To further protect label privacy, the advertiser may choose to implement differential privacy (DP) \cite{abadi2016deep, dwork2006differential}. There are two primary methods to add DP noise for label protection. The first method involves applying Gaussian or Laplace noise to the aggregated gradients. The second method employs label DP \cite{ghazi2021deep, yang2022dp, sun2022label}, where under a binary classification setting, some labels are randomly flipped to their opposite value based on a probability that is determined by the privacy budget $\epsilon$. In this paper, we focus on label DP and leave the application of noise to aggregated gradients for future exploration.

Training directly on such perturbed labels will bias conversion probability and make the CVR model miscalibrated given conversion labels are highly class-imbalanced. A simple and effective remedy is to factorized the label transition probability into loss function to debias the prediction. Denote $q_{\epsilon} = \frac{e^\epsilon}{e^\epsilon + 1}$ being the probability to keep the original label and  $l(p_i)$ being the original loss function on prediction $p_i$, the de-biased loss can be written as 
$l_{deb}(p_i) = l(p_i q_{\epsilon} + (1-p_i) (1-q_{\epsilon}))$.

\section{Experiment and Results}
\label{sec:experiments}

We conducted all experiments on internal industrial datasets, which contains more than $10$ B entries and approximately $300$ distinct features. The main predictive task is click through CVR and there are five auxiliary tasks such as click through rate (CTR) to help the main task prediction. We use ROC-AUC as the metric to evaluate model predictive performance offline as we found it correlates well with online business metrics in our settings. Table \ref{tab:lora_results} lists the offline ROC-AUC performance of LoRA fine tuning. From it, we can see that with only fine tuning $2$ millions parameters (about $1\%$ of fine tuning all parameters), we can achieve a high ROC-AUC gain (about $95\%$ of the gain from fine tuning all parameters). This is a acceptable tradeoff considering it would reduce the training time from $2.5$ days to less than $1$ hour on $1$ day's data for some big advertisers with $10$ \textit{Gbps} network bandwidth. Even with as few as $0.06$ million parameters, we can still achieve a $+3\%$ ROC-AUC gain. 

In our experiments, adding label DP will cause drop in ROC-AUC ($-2.1\%$ for $\epsilon=5$ and $-8.3\%$ for $\epsilon=3$) and  calibration issue ($1.4$ for $\epsilon=5$ and $4.0$ for $\epsilon=3$) without de-bias. With the de-bias approach, the predictions are fully calibrated ($1.0$), and the ROC-AUC drop is around $0.2\%$ - $0.5\%$ for $\epsilon$ ranging from $5$ to $3$, which is acceptable in production. For gradient compression, we tested applying QSGD \cite{alistarh2017qsgd} compression and BF16 quantization to per-sample gradients, these techniques can provide up to 4x compression rate with ROC-AUC drop around $0.6\%$ and we believe designing compression techniques tailored for per-sample gradients can further mitigate the performance loss (Table ~\ref{tab:label_dp_results}). 


Based on our online A/B experiments for conversion ads, we expect that adopting this technique could lead to a reduction of more than $10\%$ in the cost per action (CPA) for advertisers who currently lack conversion data in CVR model. This improvement is significant as it helps lower advertiser costs and increases long-term platform revenue.

\begin{table}
\centering
\caption{Results of LoRA adapter fine tuning on CVR model. Baseline is without conversion data.}
\begin{tabular}{p{0.19\textwidth}|p{0.08\textwidth}|p{0.13\textwidth}} \hline
Model setting & ROC-AUC diff (\%) & \# trainable params (millions) \\\hline
Baseline & 0 & 200 \\\hline
Fine tune all params & +3.63 & 200 \\\hline
Fine tune LoRA (rank 64) & +3.48 & 4 \\\hline
Fine tune LoRA (rank 32) & +3.44 & 2 \\\hline
Fine tune LoRA (rank 1) & +3.02 & 0.06 \\\hline
\end{tabular}
\label{tab:lora_results}
\end{table}

\begin{table}[h]
\centering
\caption{Results of Label DP and gradient compression. Baseline is no privacy constraint.}
\begin{tabular}{p{0.17\textwidth}|p{0.13\textwidth} |p{0.1\textwidth}} \hline
Model setting & ROC-AUC diff(\%) & Calibration \\\hline
Baseline & 0 & 1.0 \\\hline
No debias, $\epsilon$=3 & -8.3 & 4.0 \\\hline
No debias, $\epsilon$=5 & -2.1 & 1.4 \\\hline
Debias, $\epsilon$=3 & -0.5 & 1.0 \\\hline
Debias, $\epsilon$=5 & -0.2 & 1.0 \\\hline
QSGD, 8bit & -0.61 & 1.0 \\\hline
BF16, 16bit & -0.67 & 1.0 \\\hline
\end{tabular}
\label{tab:label_dp_results}
\end{table}
\section{Conclusion }
\label{sec:conclustion}

In conclusion, our proposed model training framework addresses privacy and efficiency challenges by using batch-level aggregated gradients, adapter-based fine-tuning methods, and label differential privacy with de-biasing techniques. Experimental results on real-world datasets indicate that our framework maintains competitive performance while adhering to privacy requirements, setting a new benchmark for privacy-preserving, high-performance CVR prediction in the digital advertising industry.

\begin{acks}
The authors would like to thank Aayush Mudgal, Andy Kimbrough, Ang Xu, Jaewon Yang, Joey Wang, Stephanie deWet, Susan Walker, Xiaofang Chen, Yingwei Li, Zhifang Liu for their valuable discussion and paper review.
\end{acks}


\appendix
\clearpage

\setcounter{figure}{0}
\renewcommand\thefigure{A\arabic{figure}} 

\section{Appendix}
\subsection{CVR model architecture}
\begin{figure}[H]
\includegraphics[width=0.75\linewidth]{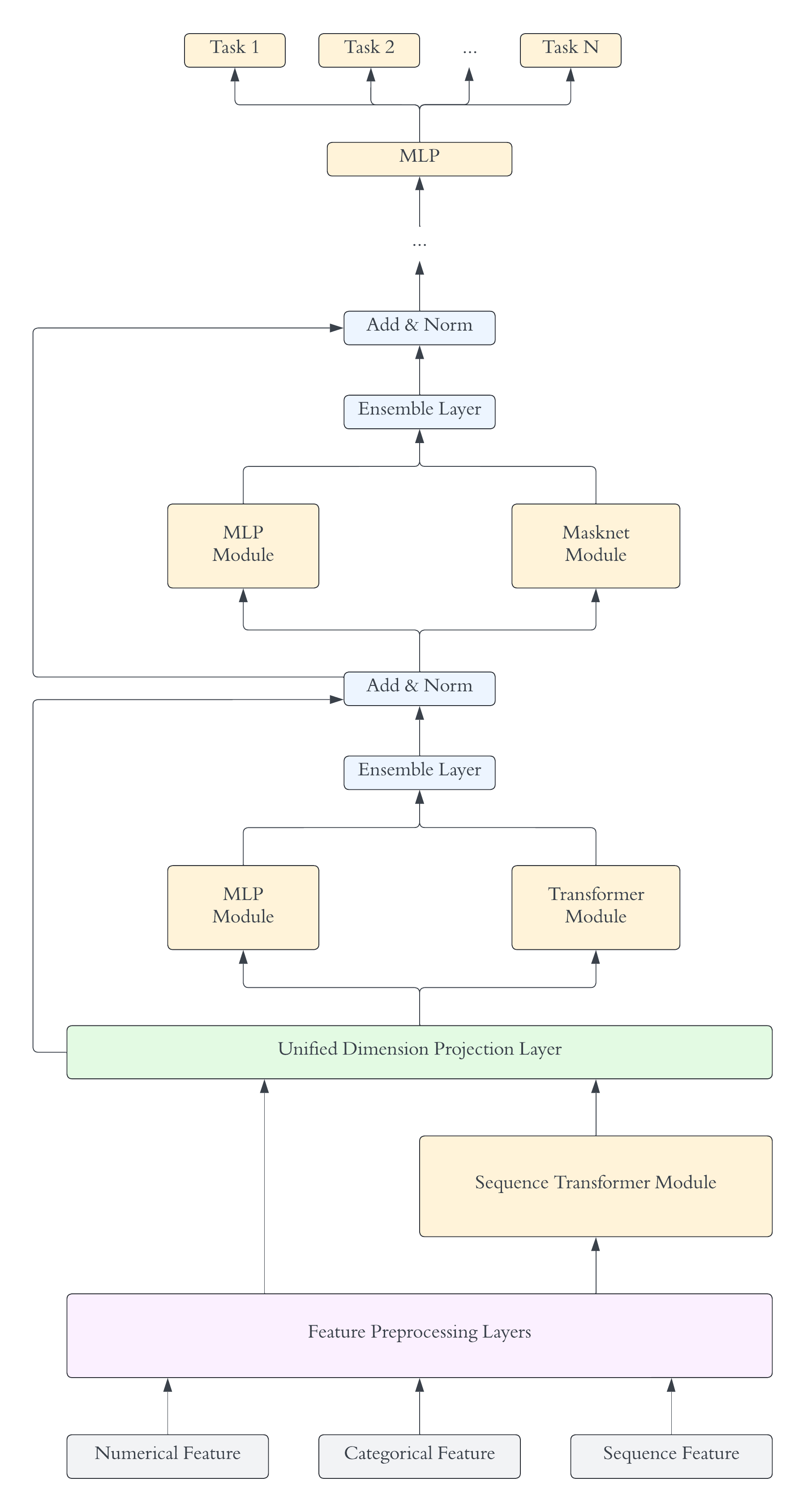}
\caption{CVR model architecture.}
\label{fig:conversion_model_arch}
\end{figure}

{\ref{fig:conversion_model_arch}} shows our CVR model architecture. We use transformers to consume the sequence features. The output vectors enter into feature interaction modules to cross with the other feature (numerical and categorical) projected vectors. Our feature interaction modules comprise of stacked layers where each layer contains heterogeneous feature crossing modules to fully capture bit-wise and vector-wise feature interactions. The final predictions have a classical multi-task learning setup. 

\subsection{Adapter based fine tuning}
Adding gated LoRA adapter layers to the CVR model: the selected layers include the QKV layer in multi-head attention in the sequence transformers, the linear layers in the feature interaction modules, and the linear layers in the multi-task towers. It has the following benefits: (1) It is parameter efficient. We only need to tune ~1-2\% parameters relative to the full model parameters; (2) It is flexible. We can add LoRA layers to both low level and high level layers in the pretrained model; (3) It is gated and we can use this unified model to serve all advertisers. During serving, only the ads candidates from the sensitive advertiser that we fine tuned on will go through LoRA layer; the advertisers in the pretraining data domain still use the original model parameters and is not impacted; (4) It is scalable to multiple advertisers. We just need to add more LoRA layers horizontally for each core module; each LoRA layer serves one advertiser. The model training can happen in parallel and independently even if the advertisers use different clean room providers.

\subsection{Label DP experiment results}
\textbf{Reduce Calibration Error}. Suppose  $p$ is the original positive probability,  $n_+$ is the number of positive instances, and $ n_{-}$ is the number of negative instances in the original dataset. Then, $p = \frac{n_+}{n_+ + n_{-}} = \frac{1}{1+\frac{n_{-}}{n_+}}$. Suppose the flipping probability is $1 - q_{\epsilon}$. After flipping, the new positive ratio is $p' = \frac{n_+ q_{\epsilon} + n_{-} (1 - q_{\epsilon})}{n_+ + n_{-}} = \frac{q_{\epsilon} + \frac{n_{-}}{n_+}(1-q_{\epsilon})}{1+\frac{n_{-}}{n_+}}$. Hence $p'=pq_{\epsilon}+(1-p)(1-q_{\epsilon})$. 
During model training, we can use the de-biasing function in our loss function $l_{deb}(p) =l(p') = l(p q_{\epsilon} + (1-p) (1-q_{\epsilon}))$ to adjust the predictions. At model serving time, we use the model output directly as our prediction, which is close to the original distribution. We demonstrated the effectiveness of the debias function as shown in Table A1.

Calibration ratio used in our experiments is calculated as $\sum_i^n \frac{p_i}{\frac{n_+}{n_+ + n_{-}}}$ where $p_i$ is the predicted probability for sample $i$. If the ratio is above 1, it means our model is over-forecasting. If the ratio is smaller than 1, it indicates that our model is under-forecasting. A perfect ratio would be 1.

\setcounter{table}{0}
\renewcommand\thetable{A\arabic{table}}

\end{document}